\documentclass{article}
\usepackage{ijcai25}

\usepackage{times}
\usepackage{amssymb}
\usepackage{soul}
\usepackage{url}
\usepackage[hidelinks]{hyperref}
\usepackage[utf8]{inputenc}
\usepackage[small]{caption}
\usepackage{graphicx}
\usepackage{amsmath}
\usepackage{amsthm}
\usepackage{booktabs}
\usepackage{algorithm}
\usepackage{algorithmic}
\usepackage[switch]{lineno}

\title{A Comprehensive Survey on Visual Concept Mining in Text-to-image Diffusion Models}

\author{
Ziqiang Li$^{1}$
\and
Jun Li$^{1}$\and
Lizhi Xiong$^{1}$\and
Zhangjie Fu$^{1}$\And
Zechao Li$^{2}$\\
\affiliations
$^1$School of Computer Science, Nanjing University of Information Science and Technology\\
$^2$School of Computer Science, Nanjing University of Science and Technology\\
\emails
iceli@mail.ustc.edu.cn, \{liwenbaisysy,lzxiong16\}@163.com, fzj@nuist.edu.cn, zechao.li@njust.edu.cn
}

\begin{document}

\maketitle

\begin{abstract}
    Text-to-image diffusion models have made significant advancements in generating high-quality, diverse images from text prompts. However, the inherent limitations of textual signals often prevent these models from fully capturing specific concepts, thereby reducing their controllability. To address this issue, several approaches have incorporated personalization techniques, utilizing reference images to mine visual concept representations that complement textual inputs and enhance the controllability of text-to-image diffusion models. Despite these advances, a comprehensive, systematic exploration of visual concept mining remains limited. In this paper, we categorize existing research into four key areas: Concept Learning, Concept Erasing, Concept Decomposition, and Concept Combination. This classification provides valuable insights into the foundational principles of Visual Concept Mining (VCM) techniques. Additionally, we identify key challenges and propose future research directions to propel this important and interesting field forward.
\end{abstract}

\section{Introduction}
\label{sec:intro}

Recently, text-to-image diffusion models (T2I-DM) \cite{rombach2022high,saharia2022photorealistic} have demonstrated performance on par with human experts in many scenarios, particularly in content creation. These models utilize text inputs to generate visually compelling images that align with textual descriptions. However, the abstractness, subjectivity, and ambiguity inherent in language often impede the precise depiction of specific object concepts, resulting in reduced controllability \cite{ruiz2023dreambooth,gal2022image} during the generation process. Efficiently and accurately mining visual concepts \cite{ruiz2023dreambooth,gal2022image,wu2025infinite} from reference images to supplement textual inputs holds promise for improving the controllability of image generation. This capability greatly enhances applications such as personalized generation across diverse scenarios. Given its customizable nature, this area has garnered significant public interest and attracted considerable research attention over the past three years.

\begin{figure*}
\setlength{\abovecaptionskip}{0.1cm}
    \setlength{\belowcaptionskip}{-0.3cm}
	\centering
	\includegraphics[scale=0.25]{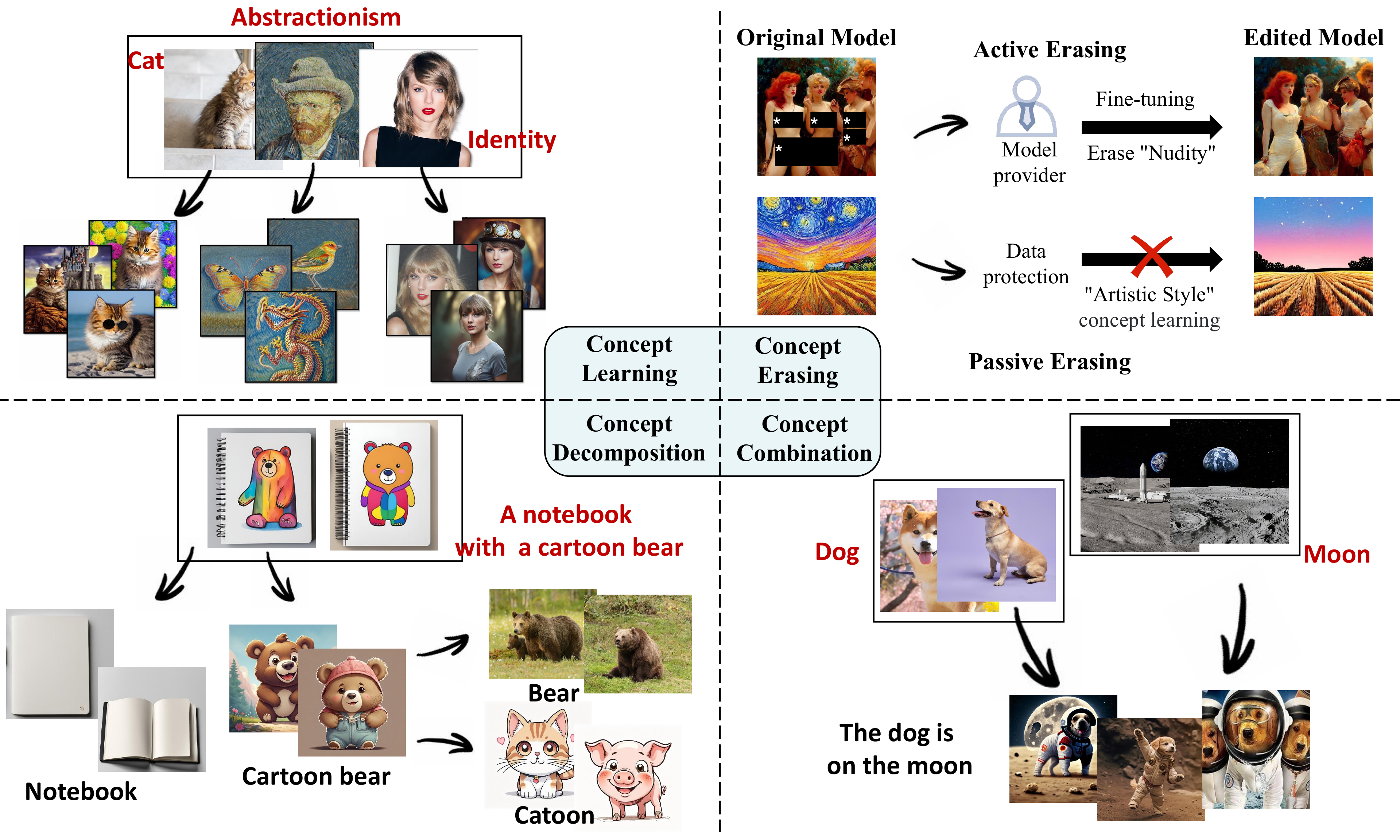}
	\caption{Visualization of several Visual Concept Mining tasks in text-to-image diffusion models.}
	\label{FIG:flow}
\end{figure*}

Recent studies \cite{li2023photomaker,wu2025infinite,vinker2023concept,wang2024instantid,ye2023ip} have widely adopted personalization techniques to extract visual concepts from various reference images. However, there is no comprehensive summary of this research direction, and its potential applications remain underexplored. As a result, there is an urgent need for systematic investigations to guide and inspire future research efforts. To address this gap, we present a thorough survey and propose a detailed taxonomy of Visual Concept Mining (VCM) in T2I-DM. VCM involves capturing key visual concepts from reference images, sometimes with as few as a single reference. Based on the different operational modes of visual concepts, we categorize existing approaches into four main types: concept learning \cite{ruiz2023dreambooth}, concept erasing \cite{gandikota2023erasing}, concept decomposition \cite{vinker2023concept}, and concept combination \cite{lee2023language}. Specifically, as illustrated in Fig. \ref{FIG:flow}, \textit{Concept Learning} introduces new individual concepts from reference images for T2I-DM, while \textit{Concept Erasing} uses reference images to remove previously learned concepts from pre-trained models. \textit{Concept Decomposition} breaks down a concept into its constituent aspects, forming a hierarchy of subconcepts, whereas \textit{Concept Combination} synthesizes learned concepts to generate novel ones. This field is rapidly evolving and holds significant value in the research community. Our paper provides a comprehensive overview of key methodologies, applications, and challenges in this area.

\noindent\textbf{Goals of our survey.} We aim to (i) develop the first thorough survey and propose a detailed  taxonomy
for VCM in text-to-image diffusion model; 
 (ii) identify open questions and challenges that could stimulate further research in VCM tasks.

\noindent\textbf{The differences between our survey and others.} 
Several survey studies have explored the text-to-image diffusion domain, covering topics such as diffusion model theories and architectures \cite{yang2023diffusion}, text-driven image editing \cite{zhan2023multimodal}, text-to-image generation \cite{zhang2023text}, text-to-video generation \cite{xing2023survey}, and text-to-3D generation \cite{li2023generative}. However, these surveys typically provide only a general overview of content generation. Additionally, the survey by \cite{cao2024controllable} offers a comprehensive review of the integration and impact of novel conditions, such as sketch, identity, style, and layout, in T2I-DM. While some of these conditions are introduced through visual concept learning, a detailed and thorough analysis of visual concept mining is lacking. The closest survey to ours is \cite{zhang2024survey}, which focuses specifically on visual concept learning but does not address other mining techniques. In contrast, our study presents a more comprehensive and in-depth survey, covering all aspects of visual concept mining, including concept learning, concept erasing, concept decomposition, and concept combination.

\section{Preliminaries}

\subsection{Denoising Diffusion Probabilistic Models (DDPMs)}
DDPMs, proposed in 2020 \cite{ho2020denoising}, have quickly become the most popular and advanced generative models. DDPMs consist of two processes: the forward process and the reverse process:

\noindent\textbf{Forward Process.} 
During the forward process, random Gaussian noises $\left(\epsilon_t\right)_{t=1}^T \sim \mathcal{N}(\mathbf{0}, \mathbf{I})$ are incrementally added to the training sample $x_0$ over a Markov chain with $T$ steps, producing a series of noisy samples $\left(\mathbf{x}_t\right)_{t=1}^T$ as follows:
\begin{equation}
\mathbf{x}_t=\sqrt{\alpha_t} \mathbf{x}_{t-1}+\sqrt{1-\alpha_t} \epsilon_t, 1 \leq t \leq T, 
\label{eq:1}
\end{equation}
where $\alpha_t$ is the variance of the Gaussian noises $\epsilon_t$. According to the additivity of the Gaussian, Eq. \ref{eq:1} has been rewritten as:
\begin{equation}
\mathbf{x}_t=\sqrt{\bar\alpha_t} \mathbf{x}_0+\sqrt{1-\bar\alpha_t} \epsilon_0,
\label{eq:2}
\end{equation}
where $\bar\alpha_t=\Pi^t_1\alpha_t$.
    
\noindent\textbf{Reverse Process.}
During the reverse process of the DDPM, various noisy samples are progressively denoised to recover the original samples. A neural network $f$ parameterized on $\theta$ has been trained to learn an estimated distribution $p_\theta(\mathbf{x}_0|\mathbf{x}_t)$. For a given random step $t$, the object of the denoised network $f(\cdot)$  is computed by minimizing the difference between the ground truth $\mathbf{x}_0$ and the estimation $\hat{\mathbf{x}}_0$:
\begin{equation}
L_{rec}  =\mathbb{E}\left[w_t\left\|\hat{\mathbf{x}}_0-\mathbf{x}_0\right\|_2^2\right]
 =\mathbb{E}\left[w_t\left\|f_\theta\left(\mathbf{x}_t, t\right)-\mathbf{x}_0\right\|_2^2\right],
 \label{eq:3}
\end{equation}
where $w_t$ represents a time-step dependent weight.
\subsection{Text-to-image Diffusion Models}
There are several pivotal and widely utilized text-to-image foundational models, such as GLIDE \cite{nichol2021glide}, Stable Diffusion (SD) \cite{rombach2022high}, and Imagen \cite{saharia2022photorealistic}. Current T2I-DM typically follow a two-step process: encoding the text prompt using pre-trained text encoders such as CLIP or T5, and then utilizing the resulting text embedding as a condition for generating corresponding images through the diffusion process. 
Specifically, text prompts $c$ are included as an additional condition to guide the reverse process. The objective of the denoising network $f(\cdot)$ can be written as:
\begin{equation}
L_{rec}
 =\mathbb{E}\left[w_t\left\|f_\theta\left(\mathbf{x}_t, t,c\right)-\mathbf{x}_0\right\|_2^2\right].
 \label{eq:4}
\end{equation}

Notably, the widely adopted Stable Diffusion \cite{rombach2022high} distinguishes itself by executing the diffusion process in latent space instead of the original pixel space, resulting in significant reductions in computation and time costs. Consequently, Stable Diffusion is the most popular open-source model, and nearly all concept mining methods build upon it. The fundamental unit of the Stable Diffusion model comprises ResBlock, self-attention layer, and cross-attention layer. The attention mechanism is represented as:
\begin{equation}
    \text{Attn}(Q,K,V)=\text{Softmax}(\frac{QK^T}{\sqrt{d}})V,
\end{equation}
where $Q$ denotes the query feature projected from the spatial features generated by the preceding ResBlock, 
$K$ and $V$ represent the key and value features. These features are projected either from the same spatial features as the query feature (in self-attention) or from the text embedding extracted from the text prompt (in cross-attention).

\section{Concept Learning}
\begin{table*}[t!]
\caption{Comparison of Three Classical Methods for Concept Learning.
	\label{Tab:Comparison_three}}
\centering
\resizebox{1\textwidth}{!}{%
\begin{tabular}{l|cccccc}
\hline 
Methods&Category & Pre-training & Inference Time & Optimized Parameters & Concept Fidelity & Semantic Consistency\\
\hline
Textual Inversion \cite{gal2022image}&Word Inversion-based Method & No & $\sim$ 20 min & Token embedding & Low & High\\
DreamBooth \cite{ruiz2023dreambooth}&Weight Optimized-based Method & No & $\sim$ 6 min & Diffusion weight & High & Low\\
ELITE \cite{wei2023elite}&Tuning-free Method & Yes & $\sim$ 4 sec & Image encoder & Moderate & Moderate\\
\hline
\end{tabular}%
}
\end{table*}
Identifying and extracting diverse visual concepts from images is a central goal in machine intelligence \cite{lee2023language}. In T2I-DM, concept learning involves extracting concepts from reference images and integrating them with a text prompt to generate customized images. These concepts can also be applied to other images for editing or recombined across instances to create novel images. The primary focus is on combining learned concepts with text prompts to generate customized images, with core evaluation criteria being concept fidelity to the reference image and semantic consistency with the text prompt. As illustrated in Fig. \ref{FIG:flow}, concept learning spans a wide range of areas, including objects, styles, identities, appearances, actions, and relationships. While these tasks share common elements, they differ in several key aspects. To this end, we first provide a systematic analysis of general concept learning from two perspectives: tuning-based and tuning-free methods. We then highlight the key differences in the learning processes across various concept types. Finally, we assert that the fusion of text and image information is a critical technology in the concept learning process, offering a comprehensive summary of this technology.
\subsection{Tuning-based Methods}
\label{sec:tune-based}
Tuning-based methods adopt reference images of the same concept to fine-tune the T2I DM or its learnable embeddings.
According to different optimized parameters, Tuning-based methods can be further categorized into \textit{Word Inversion-based methods} and \textit{Weight optimized-based methods}.

\noindent\textbf{Word Inversion-based methods.}
Word inversion-based methods, such as Textual Inversion \cite{gal2022image}, introduce a learnable token into the tokenizer to represent a subject concept. These methods optimize the token by reconstructing references from noisy inputs, maintaining high semantic consistency due to the unaltered U-Net structure. However, they compress complex visual features into a single token, which may result in prolonged convergence times and a reduction in concept fidelity. To mitigate these issues, methods such as $P+$ \cite{voynov2023p+} employ multiple token embeddings corresponding to different U-Net layers, while other approaches \cite{wang2023hifi} expand the conditioning space across various denoising timesteps, treating each as a distinct generation stage. Unlike these approaches, which typically learn a single token embedding or multiple embeddings along specific dimensions (e.g., layers or timesteps), methods like MATTE \cite{agarwal2023image} and NeTI \cite{alaluf2023neural} simultaneously consider different U-Net layers and denoising timesteps to achieve improved attribute control. Additionally, studies such as DreamArtist \cite{dong2022dreamartist}, DETEX \cite{cai2024decoupled}, and CatVersion \cite{zhao2023catversion} increase the number of tokens along the semantic dimension. These methods enable more accurate and robust concept learning by designing loss functions that allow different tokens to capture distinct semantic properties of reference images.

In contrast to directly optimizing token embeddings, several studies \cite{gal2023encoder,arar2023domain,chen2023disenbooth,achlioptas2023stellar} have trained word-embedding encoders to predict new codes in the diffusion model’s embedding space, effectively describing the input concept. These approaches offer a significant reduction in training time compared to previous methods.
In summary, word inversion-based methods introduce learnable tokens into the tokenizer to represent the concept. These methods preserve high semantic consistency by avoiding updates to the U-Net of the diffusion model. However, they rely on compressing image features into the text embedding space, which results in diminished identity fidelity.

\noindent\textbf{Weight optimized-based methods.}
Weight-optimized methods have evolved to fine-tune entire model weights, moving beyond token embeddings to better capture complex concepts. DreamBooth \cite{ruiz2023dreambooth} exemplifies this approach by fine-tuning all model parameters and introducing rare tokens (e.g., "[v*]") as unique identifiers for subjects. To mitigate overfitting and maintain diversity in generated outputs, it employs prior preservation loss. However, fine-tuning the entire model for each new concept incurs significant storage and computational costs.
To address these limitations, recent methods \cite{kumari2023multi,gal2023encoder,arar2023domain,chen2023disenbooth} focus on optimizing only key parameters, significantly reducing the time and storage required. For example, Custom Diffusion \cite{kumari2023multi} identifies cross-attention layers as the most critical for fine-tuning, despite accounting for just 5\% of the model's parameters. By restricting updates to the key and value mappings within these layers, it achieves efficient and memory-friendly fine-tuning. DETEX \cite{cai2024decoupled} refines this approach further, optimizing only the key and value mappings in cross-attention layers, achieving even greater efficiency.

Additionally, some methods \cite{arar2023domain,chen2023disenbooth,xiang2023closer} leverage the concept of Adapter \cite{houlsby2019parameter}, LoRA \cite{hu2021lora} or its variants to conduct parameter-efficient fine-tuning. Specifically, Adapter-based methods \cite{xiang2023closer,gal2023encoder} inserts adapters (small trainable modules) into the U-Net architecture of the diffusion model while keeping the majority of the model's parameters frozen. The adapters are designed to handle specific tasks with a minimal number of extra parameters (about 0.75\% of the total parameters). LoRA-based methods \cite{achlioptas2023stellar,xiao2023comcat} inserts a low-rank decomposition into pre-trained weight matrices in the U-Net, where each matrix $ W_0 \in \mathbb{R}^{d \times k}$ is decomposed as  $ W_0 \longleftarrow W_0+BA$, with $B \in \mathbb{R}^{d \times r}$, $A \in \mathbb{R}^{r \times K}$, where $r \ll min(d,k)$.  During fine-tuning, only $B$ and $A$ are learnable, while $W_0$ remains fixed, significantly reducing the number of trainable parameters from $d \times k$ to $(d + k) \times r$.



\subsection{Tuning-free Methods}
\label{sec:tune-free}
Tuning-based methods typically require extensive fine-tuning on specific concept images, a process that is not only time-consuming but also prone to limited generalization. To mitigate these challenges, recent research has explored alternative concept learning approaches that reduce the need for extensive tuning. These approaches can be broadly categorized into \textit{inversion-based} and \textit{encoder-based} methods.


\noindent\textbf{Inversion-based methods}
Inversion-based methods extract latent representations of reference images through DDIM inversion \cite{song2021denoising}. However, without additional training, these methods are limited to learning style concepts and struggle with more complex ones, such as objects or actions. Prior work \cite{tumanyan2023plug} demonstrates that attention maps encode spatial layouts, while key and value components govern content. To achieve style-content separation, DDIM inversion refines style-related attention layers, while content-related layers follow a text-guided forward pass. StyleID \cite{chung2024style} enhances this by replacing content keys and values in self-attention layers with those from a style image, particularly in late decoder stages, where textures are formed. While effective for texture transfer, this approach can disrupt content and color harmony. To mitigate this, StyleID introduces attention temperature scaling, which adjusts attention sharpness to preserve structure. Visual-Style Prompt \cite{jeong2024visual} further improves style transfer by selectively modifying upblock layers, preventing content leakage. StyleAlign \cite{hertz2023style} boosts consistency by sharing self-attention layers between reference and target images, using Adaptive Instance Normalization (AdaIN) to align styles while reducing content distortion.

\noindent\textbf{Encoder-based methods}
Inversion-based methods eliminate the need for optimization but often sacrifice fine-grained details, such as texture and color, leading to reduced concept fidelity in generated images. Furthermore, these methods can be time-consuming, limiting their practicality. Encoder-based methods \cite{qi2024deadiff,wang2023styleadapter,wang2024instantstyle,wang2024instantstyleplus,han2024stylebooth} address these challenges by pre-training image encoders on diverse datasets, enabling efficient concept learning from reference images. These approaches use specialized encoders, such as those in \cite{ye2023ip,han2024stylebooth}, which leverage Transformers with CLIP embeddings to process image features and integrate them into U-Net cross-attention layers, extracting both content and concept representations. For example, IP-Adapter \cite{ye2023ip} integrates CLIP image features into diffusion models through decoupled cross-attention, reducing textual ambiguity and improving visual fidelity. Similarly, \cite{arar2023domain} utilizes CLIP’s visual encoder and StableDiffusion’s U-Net to extract semantic and spatial features, which are then fused and processed with contrastive learning to align embeddings with the target concept.



Focusing on specific concept types like style and identity has led to specialized methods that, while utilizing pre-trained image encoders, are tailored to specific needs. StyleAdapter \cite{wang2023styleadapter} enhances style decoupling and learning efficiency with three strategies: \textit{Shuffling}, which limits concept overlap by reducing cropping; \textit{Class Embedding Suppression}, which removes class embeddings to mitigate concept entanglement; and \textit{Multiple References}, which uses multiple reference images to improve generalization across concept variations.
For identity preservation in concept learning, InstantID \cite{wang2024instantid} ensures zero-shot identity consistency by employing an identity encoder and a decoupled cross-attention mechanism. This approach extracts identity features from a reference face and combines them with textual prompts, preserving facial fidelity while allowing flexible style control. Infinite-ID \cite{wu2025infinite} advances this with the ID-Semantics Decoupling Paradigm, which separates identity and semantic information. During training, it captures identity through an image cross-attention module and merges both identity and text during inference using a hybrid attention mechanism.

\subsection{Summary}
\label{sec:learning_summary}
\noindent\textbf{Comparison of tuning-based and tuning-free methods.} As illustrated in Table \ref{Tab:Comparison_three}, tuning-based methods (e.g., DreamBooth, Textual Inversion, NeTi, CelebBasis) achieve high concept fidelity by fine-tuning models or embeddings with reference images. However, they are computationally expensive, often requiring several minutes to learn a single concept \cite{kumari2023multi}, which limits their scalability. In contrast, tuning-free methods leverage pre-trained image encoders or hyper-networks trained on large datasets, enabling efficient concept learning and customized generation with significantly lower computational overhead.

\noindent\textbf{Differences in the learning of various concept types.}
The field of visual concept learning is evolving to encompass both direct visual attributes, such as identity, style, and appearance, as well as more abstract elements like actions and relationships. Identity learning focuses on preserving unique features that define individuals, while style learning emphasizes artistic elements such as color palettes and brushstroke patterns. Both tasks require models to decouple content from distinctive traits, enabling applications like identity verification and style transfer.
In contrast to identity and style learning, other types of concept learning focus on different aspects. Appearance concept learning targets low-level visual features such as color, texture, and shape, which are essential for tasks like object recognition and image generation. Action learning, on the other hand, deals with dynamic, time-sequential patterns, requiring the separation of motion cues from static attributes. Relationship learning explores interactions between entities, emphasizing compositional structure and semantic context. These concepts vary in complexity, ranging from physical properties to dynamic behaviors and semantic abstractions, which correspond to different parts of speech in textual representations, such as nouns, adjectives, and prepositions.
While the underlying methods for these tasks share similarities, it remains important to design regularization tailored to different conceptual parts of speech. For example, ReVersion \cite{huang2023reversion} introduces a preposition prior that guides the relation prompt towards regions in the text embedding space rich in prepositional meanings, effectively enhancing relational concept learning.
\noindent\textbf{The fusion technologies of text and image information.}
An important factor influencing the performance of concept learning is how textual information is integrated with visual concepts. These integration techniques can be broadly categorized into three approaches. 
(i) Some approaches, such as PhotoMaker \cite{li2023photomaker}, combine identity embeddings from reference images with class embeddings (e.g., "man" or "woman") within the text space. However, this approach may result in compression of visual concepts, potentially affecting their fidelity.
(ii) Other studies \cite{ye2023ip,chen2023photoverse,wei2023elite} integrate visual concept representations directly into the U-Net of the diffusion model. For example, IP-Adapter \cite{ye2023ip} adds cross-attention layers to combine visual and textual information. While these methods blend identity and semantic details within the U-Net, they risk distorting the model's semantic space, potentially compromising semantic consistency.
(iii) To balance the concept fidelity and semantic consistency, some studies \cite{wu2025infinite,hertz2023style} propose merging attention maps to enhance feature interaction. This approach aims to preserve detailed visual characteristics while maintaining coherence with the text-driven generation process.
\section{Concept Erasing} 
Recent T2I-DMs, trained on large internet datasets, are capable of replicating a wide range of concepts, including undesirable ones, such as copyrighted or explicit content, which should be excluded \cite{ruiz2023dreambooth}. Concept erasure aims to remove these specific concepts from the model’s representation, preventing their unintended inclusion during image generation. This is essential for improving safety, reliability, and controllability, ensuring that generated images accurately reflect user instructions without being influenced by unwanted concepts. As shown in Fig. \ref{FIG:flow}, visual concept erasure techniques are generally divided into active and passive methods. Active erasure methods, implemented by model providers, involve fine-tuning the model to remove specific concepts, while passive erasure methods, employed by data protection entities, introduce perturbations to the data before uploading, thus preventing the model from learning these concepts and safeguarding copyright and privacy.

\subsection{Active Erasing Method}
Active erasure is typically implemented by the model provider, with its development following two key trends: i) progressing from the erasure of individual concepts to the simultaneous erasure of multiple concepts, and ii) evolving from erasing "display" concepts, which are easily recognized by the model, to erasing "implicit" concepts, which are harder for the model to control precisely. Erased Stable Diffusion (ESD) \cite{gandikota2023erasing} is the first method for concept erasure in T2I-DMs. It uses negative guidance to prevent the model from generating content related to a target concept and fine-tunes the model’s weights to reduce the likelihood of such content. This approach is efficient as it leverages the model’s internal knowledge to generate training samples, avoiding the need for additional data. \cite{kumari2023ablating} extend ESD by introducing a transformation that shifts the model's focus to an anchor concept through selective retraining of layers. Other methods, such as Forget-Me-Not \cite{zhang2024forget}, APL \cite{shi2024anonymization}, and DT \cite{ni2023degeneration}, aim to reduce unwanted content generation. Forget-Me-Not minimizes context embeddings related to the target concept, APL uses a prompt prefix to anonymize faces, and DT disrupts unwanted content by scrambling low-frequency signals.

MACE \cite{lu2024mace} addresses the limitation of existing methods, which struggle to erase a large number of concepts without affecting unrelated content, by enabling the erasure of up to 100 concepts. This is achieved through cross-attention refinement and LoRA fine-tuning, where each concept is assigned a dedicated LoRA module. To improve specificity, MACE incorporates importance sampling during LoRA training and a custom loss function to prevent interference and catastrophic forgetting. Similarly, \cite{lyu2024one} propose a one-dimensional adapter framework for multi-concept erasure, which includes a Semi-Permeable Membrane (SPM) for targeted erasure and Latent Anchoring to preserve the integrity of other concepts.

Furthermore, 
Geom-Erasing \cite{liu2024implicit} tackles the challenge of erasing implicit concepts like watermarks, which are not specified in prompts but still generated by the model. It uses an external classifier to detect the presence and location of these concepts, adding location tokens to the prompts to help the model learn their distribution. A loss re-weighting strategy further minimizes the impact of these regions, allowing the model to focus on unaffected areas and produce high-quality images while removing unwanted elements.

\subsection{Passive Erasing Method}
Passive erasing, often employed by data protection entities, aims to introduce perturbations into data to prevent diffusion models from learning concepts associated with protected content. The primary challenges of this approach are: (i) ensuring that the optimized perturbations are generalizable enough to disrupt a wide range of concept learning methods, and (ii) maintaining the robustness of these perturbations against potential data manipulations, such as image compression. AdvDM \cite{liang2023adversarial} is the first approach to employ adversarial examples for erasing visual concepts in T2I-DM. It generates subtle perturbations that impair the model's ability to recognize and extract style or content, thereby degrading the quality of generated images. Rooted in a theoretical framework, AdvDM generates adversarial examples by maximizing the training loss for diffusion models, iteratively perturbing images via Monte Carlo sampling to prevent the model from reproducing their features.

In terms of improving generalization, CAAT \cite{xu2024perturbing} focuses on optimizing the cross-attention layers, thereby minimizing computational overhead while enhancing the effectiveness of attacks. Anti-DreamBooth \cite{van2023anti} introduces a step-based perturbation optimization method, significantly improving the generalization of the erasing technique. IAdvDM \cite{zheng2023understanding} enhances adversarial attacks by jointly optimizing three key goals: misleading latent variables in both forward and reverse processes, as well as during model fine-tuning. By simultaneously optimizing these objectives, IAdvDM resolves conflicts between them, resulting in more robust and generalized attacks.

To improve robustness, InMark \cite{liu2024countering} leverages influence functions to identify and target pixels that significantly affect the denoising performance of diffusion models. By concentrating on these influential pixels, InMark ensures that even if images undergo modifications, such as compression, the personal semantics of the images remain protected. Additionally, Mist \cite{liang2023mist} enhances the robustness of adversarial examples for diffusion models by combining semantic and textual adversarial losses.


\subsection{Summary}

Despite advancements in concept erasure methods, several studies \cite{zhao2024can} highlight persistent vulnerabilities that undermine their effectiveness. For instance, concept inversion techniques can successfully recover erased concepts through specialized word embeddings, exposing the limitations of active erasure methods. Moreover, \cite{zhao2024can} identifies the shortcomings of perturbation-based passive erasure methods, which fail to completely eliminate target concepts while preserving image quality. These findings underscore the need for more robust and reliable approaches to strengthen the security of generative models.

\section{Concept Decomposition}
Visual concept decomposition seeks to deconstruct images into meaningful sub-concepts for enhanced understanding or further creation. These methods can be broadly categorized into token-based, embedding-based, and feature-based approaches. Token-based methods map visual concepts to specific textual tokens, while embedding-based methods leverage text embeddings to guide the decomposition process. Feature-based methods, on the other hand, focus on decomposing concepts by manipulating visual features directly. 

\subsection{Token-based Methods}
Token-based methods associate visual sub-concepts with text tokens, enabling the extraction of concepts that align with textual semantics, making it easier to edit and apply decomposed concepts. Specifically, Break-A-Scene \cite{avrahami2023break} optimizes text embeddings to reconstruct images with high fidelity, using mask diffusion and cross-attention losses to ensure that each token accurately corresponds to specific visual concepts. This allows for the clear decomposition and flexible combination of multiple concepts. DreamCreature \cite{ng2023dreamcreature} employs k-means clustering to identify sub-concepts in images and maps them to text tokens, enabling the generation of new mixed-concept images. Hi-CoDe \cite{qu2024llm} breaks images into hierarchical concept trees using GPT-4, mapping visual elements and attributes to text tokens, which are then linked to visual features through CLIP, refining the visual concepts further.




\subsection{Embedding-based Methods}
Embedding-based methods associate visual concepts with text embeddings. For example, \cite{vinker2023concept} uses pre-trained text-to-image models to refine text embeddings, enabling the decomposition of visual concepts into sub-concepts through a learned latent space. Additionally, some methods offer textual guidance to ensure consistency in this decomposition. Specifically, DCBM \cite{fang2024cross} decomposes image features into concept vectors (CDVs) using adversarial training, where embedding-based models such as VLM and LLM provide semantically grounded textual embeddings that guide the learning of these concept vectors. 

\subsection{Feature-based Methods}
Feature-based methods directly manipulate visual features, bypassing text tokens and embeddings. These methods decompose visual concepts through continuous representations such as vector embeddings and latent features, leveraging generative models or neural networks to capture the intrinsic relationships between visual elements. For instance, HiDC \cite{yang2020learning} utilizes visual embeddings and triplet loss to decompose images into "attribute" and "object" sub-concepts, focusing purely on visual features. Similarly, \cite{lee2023language} anchors concept embeddings to a pre-trained VQA, working within continuous embedding spaces to generate novel visual combinations and enhance generalization. Decomp Diffusion \cite{su2024compositional} applies diffusion models to break images into low-dimensional latent vectors, enabling reconstruction and editing independent of text tokens.

\subsection{Summary}
Token-based and embedding-based methods rely heavily on the quality of text embeddings, which can struggle with ambiguous or abstract concepts and fail to generalize to unseen data. In contrast, feature-based methods, lacking direct integration with text-based context, have difficulty capturing nuanced relationships, limiting their ability to effectively adopt decomposed  concepts for generating novel images.

\section{Concept Combination}

Concept combination involves integrating multiple concepts (\textit{e.g.}, subjects, scenes, styles) into a single image while maintaining semantic consistency. Existing methods can be broadly classified into two categories: (i) single-subject and scene/style combination, which blends attributes within a subject while ensuring coherence, and (ii) multi-subject combination, which focuses on preserving the identities and spatial relationships of multiple subjects.

\subsection{Single subject and scene/style combination}
Single-subject and scene/style combination generates images by merging a subject (e.g., a person or object) with a background or style. Unlike multi-subject combination, it primarily focuses on blending the subject naturally while preserving its uniqueness. The key challenge is ensuring that the subject and the background/style blend naturally, while maintaining the independence of each concept and preserving the integrity of the subject's features. Recent methods aim to seamlessly integrate subjects with background or stylistic features while preserving their uniqueness. Magicapture \cite{hyung2024magicapture} combines prompt learning with attention redirection loss (AR Loss), merging concepts like subject, background, and style through prompts treated as pseudo-labels. AR Loss decouples these concepts to avoid conflicts and ensure natural integration in high-resolution portraits. Custom Diffusion \cite{kumari2023multi} uses joint training and constraint optimization for multi-concept customization, merging concept data with distinct modifiers and fine-tuning concepts for consistent image generation.

\subsection{Multi-subject combination}

Multi-subject combination generates images with multiple distinct subjects, each maintaining its identity, position, and features. Challenges include identity confusion (blending of different subject features) and overfitting (loss of diversity or instruction adherence). Existing methods fall into two categories: tuning-based and tuning-free approaches.


\noindent\textbf{Tuning-based methods.} 
To avoid identity confusion and overfitting, existing methods differentiate subject identities while integrating multiple concepts. Break-a-Scene \cite{avrahami2023break} uses joint sampling and text prompts, generating specific prompts like "a photo of a dog and a cat at the beach" to combine concepts without blending identities. UniPortrait \cite{he2024uniportrait} integrates facial recognition and CLIP features via an ID embedding module, with DropToken and DropPath regularization, assigning unique identity indices to each spatial location for precise identity mapping in multi-subject generation.


\noindent\textbf{Tuning-free methods.} Tuning-free methods for multi-subject image generation focus on utilizing pre-trained models and architectural adjustments to integrate multiple subjects into a single image without the need for additional fine-tuning. These methods aim to address challenges such as preventing identity confusion, ensuring accurate spatial layouts, and avoiding overfitting. For instance, FastComposer \cite{xiao2024fastcomposer} tackles these challenges by employing cross-attention localization to supervise the placement of each subject within the image and delayed subject conditioning to mitigate overfitting, ensuring the preservation of subject uniqueness while maintaining spatial coherence.

\subsection{Summary}
Existing concept combination methods in text-to-image generation struggle to integrate multiple concepts while preserving each element’s independence and identity. For single-subject and scene/style combinations, the challenge lies in maintaining subject distinctiveness while seamlessly incorporating background or stylistic features. In multi-subject combinations, additional issues such as identity confusion, spatial coherence, and overfitting arise, often leading to blended identities and reduced output diversity as methods struggle to balance integration with individuality.

\section{Challenges and Opportunities}
Existing visual concept mining methods have achieved remarkable results. However, several challenges remain. Here, we aim to highlight these limitations and outline open questions that could inspire future research.
\begin{itemize}
	\item \textbf{Trade-off between concept fidelity and semantic consistency in concept learning.}  Tuning-based methods achieve high fidelity but may disrupt text-driven semantic consistency, while tuning-free methods are faster but sacrifice fidelity. Despite attention fusion or cross-modal alignment approaches, challenges like visual detail loss and semantic interference persist. Future work should explore dynamic fusion or hierarchical controls to better align visual concepts with text constraints, potentially through fine-grained interactions in diffusion model layers or semantic guidance modules.

	\item \textbf{A unified framework for concept learning across all categories.} Current methods rely on independent strategies for different concept types (e.g., identity, style, relationships), limiting cross-concept collaboration. A unified framework using modular designs (e.g., dynamic adapters, hierarchical encoding) and large-scale multimodal pretraining can enhance scalability and generalization. Integrating LLMs could further improve abstract concept expression and combination.

    \item \textbf{Robust, generalizable, and scalable concept erasure.}
Existing erasure methods face robustness and scalability challenges. Active techniques are vulnerable to adversarial recovery, while passive methods struggle with perturbation generalization and post-processing disruptions. MACE improves scalability but increases storage overhead. Future work should focus on lightweight erasure frameworks with adversarial training and causal reasoning-based decoupling to enhance security and scalability.
    \item \textbf{Flexible and high-fidelity decomposition and composition of visual concepts.}
Decomposition methods struggle with abstract concepts, and composition methods often lead to identity confusion or style conflicts. Orthogonal Adaptation improves independence but lacks flexibility. Future work should develop decoupled representations based on geometric or semantic structures, and optimize using diffusion priors. Incorporating commonsense constraints (e.g., LLM-guided layout generation) can improve logical coherence and visual consistency.

\end{itemize}
\section{Conclusion}
This paper presents a comprehensive survey and taxonomy of visual concept mining (VCM) in text-to-image diffusion models. We categorize existing VCM techniques into four main types based on their operational modes: concept learning, concept erasure, concept decomposition, and concept composition. Finally, we highlight key challenges and open questions that warrant further exploration in future research.
\clearpage
\small
\bibliographystyle{named}
\bibliography{ijcai25}

\end{document}